\documentclass[letterpaper]{article} 
\usepackage{aaai25}  
\usepackage{times}  
\usepackage{helvet}  
\usepackage{courier}  
\usepackage[hyphens]{url}  
\usepackage{graphicx} 
\usepackage{natbib}  
\usepackage{caption} 
\usepackage{algorithm}
\usepackage{algorithmic}
\usepackage{bm}
\usepackage{amsmath, amssymb}
\usepackage{multirow}
\usepackage{booktabs}
\usepackage{color}
\usepackage{xcolor}

\def\vb{{\bm{b}}}
\def\vf{{\bm{f}}}
\def\vm{{\bm{m}}}

\DeclareMathOperator{\sign}{sign}
\newcommand*{\tran}{^{\mkern-1.5mu\mathsf{T}}}

\usepackage{newfloat}
\usepackage{listings}
\DeclareCaptionStyle{ruled}{labelfont=normalfont,labelsep=colon,strut=off} 
\floatstyle{ruled}
\newfloat{listing}{tb}{lst}{}
\floatname{listing}{Listing}
\title{All You Need in Knowledge Distillation Is a Tailored Coordinate System}
\author{
    Junjie Zhou,
    Ke Zhu,
    Jianxin Wu\thanks{J. Wu is the corresponding author.}
}
\affiliations{
    National Key Laboratory for Novel Software Technology, Nanjing University, China\\
    School of Artificial Intelligence, Nanjing University, China\\
    zhoujj@lamda.nju.edu.cn, zhuk@lamda.nju.edu.cn, wujx2001@gmail.com
}

\begin{document}

\maketitle

\begin{abstract}
	Knowledge Distillation (KD) is essential in transferring dark knowledge from a large teacher to a small student network, such that the student can be much more efficient than the teacher but with comparable accuracy. Existing KD methods, however, rely on a large teacher trained specifically for the target task, which is both very inflexible and inefficient. In this paper, we argue that a SSL-pretrained model can effectively act as the teacher and its dark knowledge can be captured by the coordinate system or linear subspace where the features lie in. We then need only one forward pass of the teacher, and then tailor the coordinate system (TCS) for the student network. Our TCS method is teacher-free and applies to diverse architectures, works well for KD and practical few-shot learning, and allows cross-architecture distillation with large capacity gap. Experiments show that TCS achieves significantly higher accuracy than state-of-the-art KD methods, while only requiring roughly half of their training time and GPU memory costs.
\end{abstract}

\section{Introduction}

Deep learning has improved areas such as computer vision and natural language processing. However, large model size and heavy compute make it difficult to deploy them on devices with limited resources. Model compression methods, especially knowledge distillation~\cite{hinton2015distilling}, effectively compress and accelerate deep networks, as evidenced in various domains~\cite{hinton2015distilling, wang2021distilling, yang2020textbrewer}.

Knowledge Distillation (KD) has emerged as a crucial technique for enhancing the efficiency of neural networks by transferring ``dark knowledge'' from a large complex model (the teacher) to a smaller one (the student). In spite of its many successes, KD still encounters many drawbacks and limitations:
\begin{itemize}
	\item \textbf{Reliant} on a teacher trained for the specific task, which is highly costly, cumbersome and inflexible. Utilizing a \emph{self-supervised-learning (SSL) pretrained model as the teacher} is highly desired, but is rarely studied or with limited accuracy~\cite{abbasi2020compress}. Self distillation by~\citet{zhang2019your} is teacher-free, but incurs significantly higher memory and time costs.
    \item \textbf{Inflexible}. KD mostly leverages the logits as training signals, which is inflexible, e.g., for complex tasks like object detection~\cite{wang2021distilling}. Difficulties are also evident when the teacher has huge capacity but the student is a small network~\cite{mirzadeh2020improved}.
    \item \textbf{Inefficient}. Feature-based KD~\cite{tian2019contrastive,chen2021wasserstein, wang2021distilling, miles2024vkd} is more flexible than logit-based ones by mimicking final and/or intermediate features. However, both feature- and logit-based KD methods require a well-trained teacher and the teacher's forward passes, which nearly doubles the time and memory costs.
\end{itemize}

Our goal is to achieve \emph{teacher-free, flexible and efficient} knowledge distillation. We will extract knowledge from SSL pretrained models, which does \emph{not} require training for any specific task, that is, being teacher-free for KD. Two key questions remain open in the literature: 
\begin{enumerate}
 \item What is the dark knowledge inside a pretrained model? 
 \item How to efficiently adapt it for a specific task?
\end{enumerate}

Our answer to both questions is to propose a tailored coordinate system (TCS). Our key hypothesis is that \emph{the dark knowledge within a SSL pretrained model are at least partly encoded within the coordinate system} (or linear subspace) in which their features lie in. The linear subspace can be efficiently encoded via PCA (principal component analysis).

To adapt this coordinate system to a specific task and the student network, we propose to simply compute PCA of the teacher's features using \emph{one single forward pass} on that task's training data. Furthermore, the key to adapt the coordinate system is to \emph{tailor it by choosing a subset of the coordinates specifically for the target task}.

Combining both key techniques, we argue that all you need in knowledge distillation is a tailored coordination system, i.e., the proposed TCS method.

\begin{figure}
	\centering
	\includegraphics[width=1.0\columnwidth]{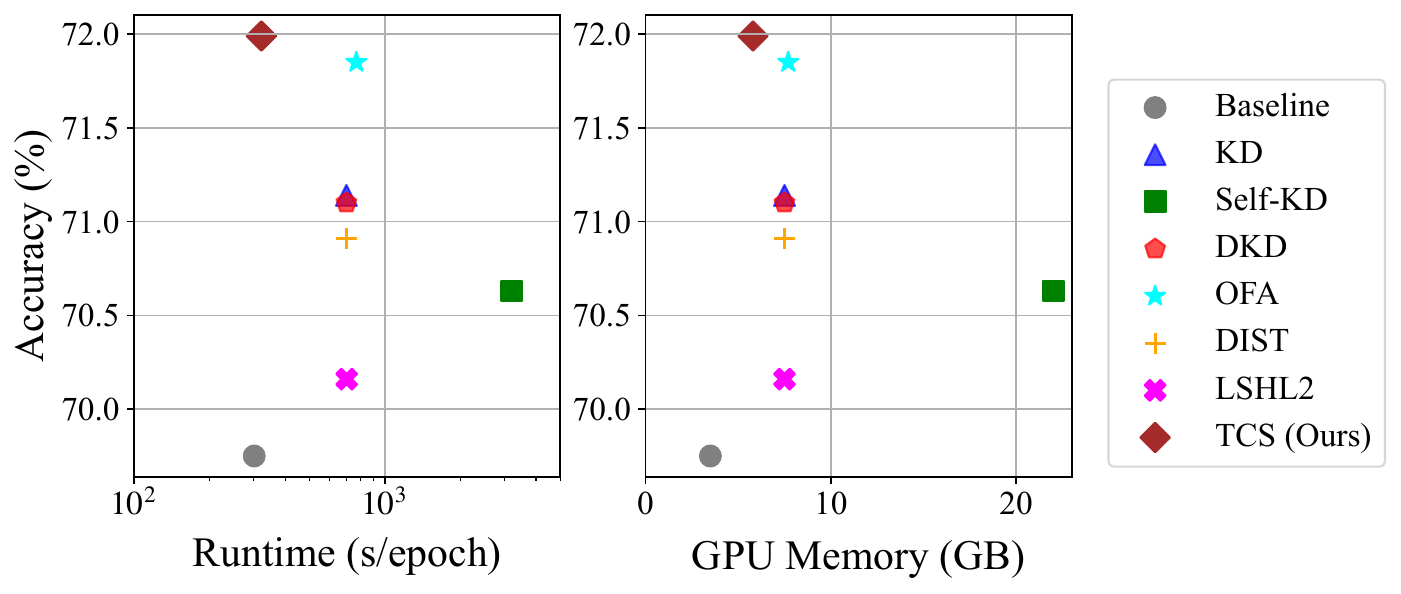} 
	\caption{Top-1 accuracy, training time and GPU memory consumption on ImageNet-1K. Our TCS achieves the fastest training, smallest GPU memory footprint and highest accuracy among KD methods. In the left figure,  x-axis is logarithmic. The teacher is Swin-Tiny and the student is ResNet18.}
	\label{fig:fig1}
\end{figure}

As shown in Figure~\ref{fig:fig1}, the proposed TCS achieves the highest accuracy with lowest memory and compute costs among KD methods. In addition, it enjoys the following benefits:
\begin{itemize}
	\item Teacher-free, i.e., effectively transfers knowledge from SSL pretrained models.
	\item Easily applicable to diverse architectures such as CNNs, Transformers, and MLPs.
	\item Spans both traditional KD and few-shot learning.
	\item Allows cross-architecture and huge capacity gap between teacher and student.
	\item Achieves almost the same training speed as training the student from scratch.
\end{itemize}

In summary, our TCS is flexible, efficient, and not reliant on task-specific teacher. Technically, our contributions are:
\begin{itemize}
    \item We argue that tailored coordinate system (TCS) is all you need to transfer dark knowledge from a task-agnostic teacher to a task-specific student network. Our TCS, of course, can also use a teacher trained for the target task in a supervised manner.
    \item Our TCS method is flexible and versatile, which accommodates diverse architectures and tasks. It is efficient in both KD training and inference in terms of speed and memory footprint, and delivers superior accuracy on both traditional KD and few-shot learning.
\end{itemize}

\section{Related Work}

Knowledge Distillation (KD) is a line of methods to transfer knowledge from large, (often) high-capacity teacher networks to smaller, more efficient student ones.

\paragraph{Knowledge Distillation} Originally proposed by~\citet{hinton2015distilling}, KD facilitates the transfer of knowledge in two primary domains: output and representation spaces. Direct logits transfer has been extensively explored~\cite{jin2019knowledge, li2020local, yuan2020revisiting, yun2020regularizing, mirzadeh2020improved, son2021densely, zhou2021rethinking, kim2021self, zhao2022decoupled, hao2024one}, along with methods focusing on feature or representation distillation~\cite{zagoruyko2016paying, huang2017like, kim2018paraphrasing, park2019relational, heo2019comprehensive, tian2019contrastive, chen2021distilling, wang2021distilling, zhu2023quantized}. The core in KD is to encourage students to emulate or replicate the predictions of teacher models. KD has been widely applied across diverse learning scenarios~\cite{Yu_2017_ICCV, chen2017learning, kd-gift, kickstart-rl}.

KD methods, however, mostly requires a teacher that is specifically trained for the target task. SSL-pretrained models are less studied in KD~\cite{abbasi2020compress}. And, KD methods nearly double the compute and GPU memory costs compared to training the student from scratch, even when excluding the costs associated with training the teacher itself. A recent method~\cite{zhang2019your} introduced self-distillation by condensing knowledge internally, thus effectively reducing the network's size. This method is teacher-free but has very large compute and GPU memory costs.

\paragraph{Practical Few-Shot Learning} Different from classical Few-Shot Learning (FSL), practical Few-Shot Learning or pFSL is tailored to better accommodate real-world scenarios~\cite{fu2024ibm2}. Traditional FSL predominantly relies on knowledge derived from a base set of labeled images, employing strategies such as meta-learning or transfer learning. Meta-learning methods train across multiple episodes, enabling rapid adaptation to new tasks with minimal data~\cite{hospedales2021meta_survey}, while transfer-learning typically involves pretraining on base classes followed by fine-tuning on novel classes~\cite{chen2019closer, mangla2020charting}.

In contrast, pFSL foregoes reliance on a predefined base set, opting instead for SSL models pretrained on large-scale datasets~\cite{fu2024ibm2}. Our distillation methodology for few-shot learning incorporates the pFSL paradigm to enhance real-world applicability and effectiveness.

\section{Method}

\begin{figure*}
	\centering
	\includegraphics[width=1.0\textwidth]{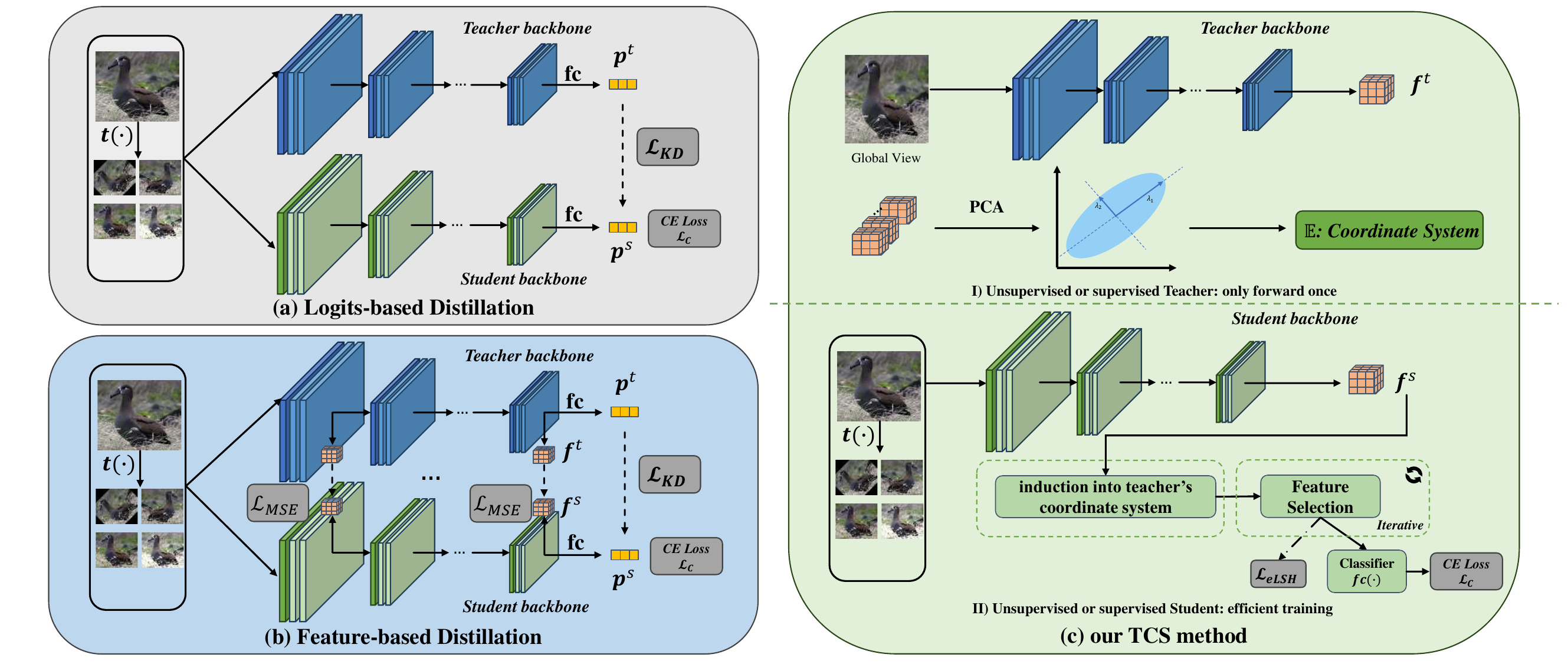}
	\caption{Illustration of existing KD methods and our TCS method. (a) logits-based distillation, where the student learns only from the final predictions of the teacher ($\mathbf{p}^t$); (b) feature-based distillation, in which the student mimics intermediate features of the teacher ($\mathbf{f}^t$), too; and (c) our TCS method, which aligns student features into the teacher's coordinate system and \emph{tailor} it to the target task by feature selection. TCS is optionally augmented by an eLSH module---an efficient unsupervised feature mimicking method. This figure is best viewed in color.}
	\label{approach}
\end{figure*}

As depicted in Figure~\ref{approach}, our TCS starts by extracting the teacher's features through the training dataset. The extracted features are used to compute the teacher's coordinate system by Principal Component Analysis (PCA). This procedure does \emph{not} involve data augmentation, which means \emph{only one} forward pass is needed and the cost involving the teacher is in fact negligible.

We then align the student's features to fit into this coordinate system. This alignment process is supported by our iterative feature selection method, which tailors the coordinate system to suit the target task. eLSH is another module we proposed by a simple modification to LSH feature mimicking of~\citet{wang2021distilling}, which further improves our TCS efficiently. 

We present our TCS method and its modules one by one.

\subsection{Preparations}

In contrast to methods that utilize intermediate features from various layers within the backbone as supervisory signals, our TCS is inspired by the ``feature mimicking'' approach suggested by~\citet{wang2021distilling}, which only mimics features from the final (penultimate) layer in the backbone. 

For an input image $x$, the teacher generates a feature vector $\vf^t \in \mathbb{R}^{D_t}$, while the student produces $\vf^s \in \mathbb{R}^{D_s}$. To reconcile the differing dimensionalities between them, a linear layer is inserted after the student backbone. It changes $\vf^s$ from $D_s$ dimensions to $D_t$ by a fully connected (FC) layer, and the transformed feature vector is $\hat{\vf}^s \in \mathbb{R}^{D_t}$. 

When the student is not trained from scratch, the linear transformation matrix $W$ in this FC layer is initialized by using the feature sets $F^t = \{\vf^t_i\}_{i=1}^N$ ($N \times D_t$) and $F^s = \{\vf^s_i\}_{i=1}^N$ ($N \times D_s$), where $N$ denotes the number of training images. $W$ is initialized as the optimal least-square solution that regresses $F^t$ from $F^s$: 
\begin{equation}
	W \leftarrow (F_s\tran F_s)^{-1}F_s\tran F_t \,.
\end{equation}

\subsection{Dark knowledge from a SSL-pretrained teacher}

There is no doubt that a SSL-pretrained model~\cite{seva,cropping} contains useful dark knowledge for various downstream tasks, as even finetuning such models with few-shot samples lead to excellent recognition and detection results~\cite{fu2024ibm2}. However, the difficulty is \emph{how to transfer such dark knowledge}?

Logit- or feature-based KD are both implicit---they mimic either the logit or feature of the teacher. But, because the SSL-pretrained model is \emph{not} trained for the target task, its logit or feature is at best suboptimal (if not harmful).

Our theoretical hypothesis and empirical finding is that the \emph{coordinate system, or the linear subspace where the teacher's features reside in}, contains enough dark knowledge for various downstream target tasks.

We are inspired by the finding of~\citet{yu2023compressing}, which states that the features of various models are low-rank and the PCA reduced lower-dimensional versions of these features contain almost the same amount of useful information as the original features. 

In other words, the few orthonormal eigenvectors that define the projection directions in PCA form a new and (semantically almost equivalent) coordinate system for the teacher's features. And, because \emph{we compute PCA using the target task's training data}, this coordinate system already incorporates some characteristics of the target task in addition to the dark knowledge from the SSL-pretrained model.

Of course, there must be significant domain gap between the general-purpose SSL-pretrained model and the target task, or, we must \emph{not} assume all PCA dimensions are useful for the target task. We further hypothesize that we can \emph{select a subset of the PCA dimensions (or coordinates)} that suit the target task. Hence, our overall hypothesis is that we can tailor (i.e., feature selection) the teacher's coordinate system defined via PCA. Hence, our proposed method is called TCS (tailored coordinate system).

We want to emphasize that even though our hypotheses assume a SSL-pretrained teacher~\cite{ssl-cvpr}, the logic also clearly applies when the teacher is trained using the target task's data. That is, our hypotheses are also suitable for traditional KD tasks.

\subsection{Induction into teachers' coordinate system}

The teacher's feature matrix, $F^t = \{\vf^t_i\}_{i=1}^N$, is $N \times D^t$ in size. The first step in PCA is to center the features by subtracting the mean of each column from its respective values:
\begin{align}
	\bm{\mu} &= \frac{1}{N} \sum_{i=1}^N \vf^t_{i}, \\
	X &= F^t - \mathbf{1}_N \bm{\mu}\tran \,,
\end{align}
where $\mathbf{1}_N$ is an $N \times 1$ vector of all ones.

PCA is then efficiently computed via the Singular Value Decomposition (SVD) of the centered matrix 
$X = U \Sigma V\tran$, where $U$ and $V$ are orthogonal matrices containing the left and right singular vectors of $X$, respectively, and $\Sigma$ is a diagonal matrix containing the singular values. The columns of $V$ (right singular vectors) represent the principal components of $X$. These principal components define the new coordinate system along which the variations of the data are maximized.

Induction of a student feature $\hat{\vf}_s$ into this coordinate system is pretty simple: by projecting $\hat{\vf}_s$ onto this coordinate system to obtain a new representation $\tilde{\vf}^s$:
\begin{equation}
\tilde{\vf}^s = (\hat{\vf}^s - \bm{\mu}) V \,.
\end{equation}

The features in $\tilde{\vf}^s$ encode knowledge from the pretrained model. As showed by the distributed representation nature of deep features, one single dimension of $\vf^t$ often does not correspond to semantic concepts well, but the combination of them can~\cite{vi:Bengio2013}. For example, the linear combination formed by the first column in $V$ (i.e., the first dimension in $\tilde{\vf}^s$) may have high correlation with the concept ``animal''. By encouraging the first dimension of $\tilde{\vf}^s=(\hat{\vf}^s - \bm{\mu}) V$ and that of $(\vf^t - \bm{\mu}) V$ to be similar, the student will benefit from this coordinate system when the target task is related to animals. Instead, classic KD method ask $\hat{\vf}^s$ and $\vf^t$ to be similar, which cannot enjoy such benefits due to the distributed representation nature.

\subsection{Iterative feature selection}

On one hand, it is also obvious that not all target tasks are related to all concepts as in the general-purpose SSL-pretrained teacher. For example, the target task may be flower recognition, which is irrelevant to animals. Many dimensions in $\tilde{\vf}^s$ may well be irrelevant or even noise in the target task. We counter this difficulty by feature selection, i.e., choosing a subset of dimensions from $\tilde{\vf}^s$ that are useful for the target task. The subset of features that are picked up forms the Tailored Coordinate System (TCS) for the student.

Since $\tilde{\vf}^s \in \mathbb{R}^{D_t}$, we introduce a trainable mask $\vm \in \mathbb{R}^{D_t}$, which is initialized as $\mathbf{1}_{D_t}$ and gradually changes some dimensions in $\vm$ to 0. This simple feature selection strategy prunes irrelevant dimensions using the target task's training data. Subsequently, the student features become
\begin{equation}
\bar{\vf}^{s} = \tilde{\vf}^s \odot \vm \,.
\end{equation}

For feature selection, we resort to the $\ell_{1}$-norm:
\begin{equation}
\mathcal{L} = \mathcal{L}_{CE} + \lambda \|\vm\|_{\ell_1} \,,
\end{equation}
where $\mathcal{L}_{CE}$ is the loss for training the student model (with cross entropy being the most frequently used one), and $\lambda$ is the hyperparameter to balance $\mathcal{L}_{CE}$ and $\|\vm\|_{\ell_1}$.

Our iterative feature selection module retains relevant dimensions by iteratively removing the influence of irrelevant ones. We accumulate the gradient with respect to $\vm$:
\begin{equation}
\Delta^{(\tau)} \leftarrow \frac{1}{n} \sum_{i=1}^{n} \nabla_{\vm} \mathcal{L}(\bm{\theta}^{(\tau)}, \vm^{(\tau)}) \,,
\end{equation}
where $n$ is the batch size, $\tau$ indexes training iterations, and $\bm{\theta}$ represents the parameters of the student model. We collect these gradient contributions over iterations, $\sum_{i=0}^{\tau} \Delta^{(i)}$, to continuously assess each dimension's relevance. The selection strategy at iteration $\tau$ is:
\begin{equation}
S^{\tau} \leftarrow {\tt TopDims}(\sum_{i=0}^{\tau} \Delta^{(i)}, r_{\tau} \cdot D_t) \,,
\end{equation}
where $r_{\tau}$ represents the target selection ratio at that iteration and $\mathtt{TopDims}(\bm{a},b)$ picks the largest $b$ dimensions from $\bm{a}$. The update rule for $\vm$ in each iteration is then:
\begin{equation}
\vm^{(\tau+1)} \leftarrow \vm^{(\tau)} \odot (1 - S^{\tau}) + \left(1 - \frac{r_{\tau}}{r}\right) \odot S^{\tau} \,.
\end{equation}
This formula ensures a smooth transition towards the final target selection ratio $r$, which is set to 0.5 in all experiments.

It is worth noting that operations after the student's backbone (linear layer to obtain $\hat{\vf}^s$, PCA to obtain $\tilde{\vf}^s$ and iterative feature selection to obtain $\bar{\vf}^{s}$) are all \emph{linear}. During inference, TCS can seamlessly integrate all these modules into the student's classifier. Hence, TCS does \emph{not incur extra parameters or computational overhead} during the student's inference.

\begin{table*}[t]
	\centering
	\small
	\setlength{\tabcolsep}{5pt}
	\begin{tabular}{lcccccccccccc}
	  \toprule
	  \multirow{2}{*}{Teacher} & \multirow{2}{*}{Params}  & \multicolumn{2}{c}{From Scratch} & \multicolumn{9}{c}{KD Methods}  \\
	  \cmidrule(lr){3-4} \cmidrule(lr){5-13}
							   &     &  T.     & S.     & Self-distill & KD     & DKD  & CRD & DIST  & OFA    & LSHL2$^\sharp$ & \textbf{TCS-}   & \textbf{TCS} \\
	  \midrule
	  ViT-S                    & 21.7M  & 92.04  & 76.72  &  80.50   & 77.26  & 78.10  & 76.60 & 76.49 & 80.15  & 80.38  & 78.23 & \textbf{81.37} \\
	  Memory (GB)                                        &                      & -      & 3.5      &  22.0     & 7.5    & 7.5    & 3.9    &  7.5 & 7.6 & 7.5 &  5.6  & 5.8     \\
	  Runtime (s/ep)                                  &                      & -      & 29.34      &  385.36    & 49.90   & 50.83   & 68.05   & 50.91 & 55.07 & 52.02  & 29.88  & 32.14    \\
	  \cmidrule(lr){1-13} 
	  ConvNeXt-T               & 27.9M                      & 88.41  & 76.72  &  80.50  & 79.35  & 80.43 & 78.87 & 77.75 & 80.91  & 80.24  & 78.35 & \textbf{81.60} \\
	  Memory (GB)                                       &                      & -      & 3.5      &   22.0    & 7.5    & 7.5        &    3.9  & 7.5   & 7.7  & 7.5    & 5.6   & 5.8     \\
	  Runtime (s/ep)                                          &                      & -      & 29.34      &  385.36    & 55.71   & 56.66   & 73.09   & 57.06  &  61.07      & 56.41  & 30.84  & 32.71    \\
	  \cmidrule(lr){1-13} 
	  Mixer-B/16               & 59.2M                       & 87.29  & 76.72  & 80.50  & 77.79  & 78.67 & 76.42                           &  76.36 & 79.39  & 79.37  & 78.80 & \textbf{81.02}  \\
	  Memory (GB)                                        &                      & -      & 3.5      &    22.0  &  7.6    & 7.6    & 4.2    & 7.6    & 8.1  &  7.6      & 5.7  & 5.8      \\
	  Runtime (s/ep)                                    &                      & -      & 29.34      &   385.36  & 71.34   & 71.79   & 88.94   & 72.37  &   77.33     &   72.44    & 30.73  & 31.16    \\
	  \bottomrule
	\end{tabular}
	\caption{Results of KD methods on CIFAR-100. The student is ResNet18, which has 11.6M parameters. Best results are shown in boldface. $^\sharp$ indicates results run by us, and other results are directly obtained from respective papers. All metrics were measured on a single NVIDIA 3090 GPU, and all experiments were conducted with the same batch size.}
	\label{tab:cifar}
\end{table*}

\subsection{Efficient feature mimicking}

On the other hand, when a teacher model trained specifically for the target task is indeed available (i.e., the teacher is no longer a SSL-pretrained one and the student is trained from scratch), we believe that all dimensions in $\vf^t$ are useful for guiding the student. In this case, feature selection is not needed, and we can use $\vf^t$ to directly guide the learning of the student.

As we showed earlier, the coordinate system defined by PCA is beneficial for such guidance. Hence, we want $\tilde{\vf}^s$ to mimic $W\tran (\vf^t-\bm{\mu})$, because they are the student and teacher features in this coordinate system, respectively.

To make these two feature vectors similar, we follow the LSH loss by~\citet{wang2021distilling}. The LSH loss efficiently hashes similar features into the same `bins', which reduces computational costs in high-dimensional data spaces:
\begin{align}
	\mathbf{h} &= \sign\left(W\tran(\vf^t-\bm{\mu})V+\vb\right) \,, \\
	\mathbf{p} &= \sigma\left(W\tran\tilde{\vf}^s+\vb\right) \,,
\end{align}
where $\sigma$ is the sigmoid function and $\mathbf{h}$ and $\mathbf{p}$ are (hard and soft) hash codes for the teacher and student, respectively. Suppose there are $M$ hash codes (i.e., $\mathbf{h},\mathbf{p}\in\mathbb{R}^M$), the loss that forces $\tilde{\vf}^s$ to mimic $(\vf^t-\bm{\mu})V$ is
\begin{equation}
 \mathcal{L}_{eLSH}=-\frac{1}{M}\sum_{j=1}^{M}\left[h_j\log p_j+(1-h_j)\log(1-p_j)\right] \,,
\end{equation}
following~\citet{wang2021distilling}. We also set all hyperparameters ($M$ and the weight to combine eLSH with $\mathcal{L}_{CE}$) as the default values in~\cite{wang2021distilling}.

However, a major difference between our eLSH loss and the LSH loss by~\citet{wang2021distilling} is that we do \emph{not} involve data augmentation in computing $\mathbf{h}$. Hence, only one forward pass is needed to compute $\mathbf{h}$, while in~\cite{wang2021distilling} $\mathbf{h}$ needs to be computed in every epoch. Hence, we name our loss as efficient LSH (eLSH).

For traditional KD, our method is also called TCS, and we refer to our method without eLSH as 'TCS-'. When the teacher is SSL-pretrained, eLSH is not used---as aforementioned, it is not suitable in that case.

It is also worth mentioning that the KL-divergence loss used for logit-distillation is \emph{not} used in our method.

\section{Experimental Results}

We assess our method across two distinct experimental setups: traditional KD, and the pFSL framework applied to few-shot scenarios. We provide essential information about our experiments' data, training and evaluation in this section. For more details, please refer to our supplementary materials.

\subsection{Experiments in traditional KD}

We experimented with CIFAR-100~\cite{krizhevsky2009learning} and ImageNet-1K~\cite{russakovsky2015imagenet}. CIFAR-100 contains 50,000 training and 10,000 test samples, each with a resolution of $32 \times 32$ pixels. ImageNet-1K is a larger dataset, comprising of 1.2 million training images and 50,000 validation images, all at a resolution of $224 \times 224$ pixels. To accommodate the input requirements of Vision Transformers (ViTs) and MLPs, which process image patches, we upsample CIFAR-100 images to $224 \times 224$ pixels for all experiments.

In our experiments, we trained the student network on CIFAR-100 using the code and hyperparameters provided by CRD~\cite{tian2019contrastive} to ensure consistent and fair comparisons. For ImageNet, we adhered to the standard PyTorch practices, which trains a model for 100 epochs as per CRD guidelines to strike a balance between computational efficiency and learning quality.

To address randomness in the hashing process within the eLSH module, we followed recommendations of~\citet{wang2021distilling} by using the average model from the final 10 epochs of training. This approach helps stabilize the model's performance by averaging out fluctuations in the late training stages.

\begin{table}
	\centering
	\small
	\setlength{\tabcolsep}{1.5pt}
	\begin{tabular}{lccccccc}
	  \toprule
	  \multirow{2}{*}{Teacher} & \multirow{2}{*}{Acc} & \multicolumn{6}{c}{KD Methods}  \\
	  \cmidrule(lr){3-8} 
									&        & KD     & DKD    & OFA    & DIST & LSHL2$^\sharp$    & \textbf{TCS} \\
	  \midrule
	  DeiT-T                    & 72.17      & 70.22  & 69.39  & 71.34  & 70.64  & 70.13 & \textbf{71.40} \\
	  Memory                                      & -            &  7.5   &  7.5    &  7.6    &   7.5   &  7.5   &   5.8    \\
	  Runtime                                     & -           &  415.33   &  419.02   &  481.28   &   419.36  & 419.69  &  328.24    \\
	  \cmidrule(lr){1-8} 
	  Swin-T                                    & 81.38   & 71.14  & 71.10  & 71.85  & 70.91 & 70.16   & \textbf{71.99} \\
	  Memory                                    & -       & 7.5   & 7.5    & 7.7    & 7.5    & 7.5  & 5.8     \\
	  Runtime                                    & -        & 701.80   & 701.73   & 768.68   & 703.29   & 703.34  & 321.27    \\
	  \cmidrule(lr){1-8} 
	  Mixer-B                                     & 76.62    & 70.89  & 69.89  & 71.38  & 70.66 & 70.18   & \textbf{71.45}  \\
	  Memory                                     & -           &  7.6    & 7.6    & 8.1   & 7.6    & 7.6   & 5.8      \\
	  Runtime                                 & -          & 869.97   & 868.95   & 937.92   & 872.72   & 877.16  & 341.88    \\
	  \bottomrule
	\end{tabular}
	\caption{Results on ImageNet-1K. ResNet18 is the student. When trained from scratch, its top-1 accuracy is 69.75\%. Self-distillation~\cite{zhang2019your} achieved an accuracy of 70.63\%, with a memory usage of 22 GB and runtime of 3207.45 seconds per epoch.}
	\label{tab:imagenet}
\end{table}

We evaluated with a variety of models with heterogeneous architectures. Specifically, we employed ResNet~\cite{he2016deep} as a representative CNN model, ViT~\cite{dosovitskiy2020image}, DeiT~\cite{touvron2021training}, Swin~\cite{liu2021swin} as ViT models, and MLP-Mixer~\cite{tolstikhin2021mlp} for MLP-based ones. In our training framework, the features are those from the penultimate layer, which precedes the final classification layer. Specifically, for Vision Transformers (ViT), this corresponds to the $\mathtt{CLS}$ token, while for CNN-like models, it refers to the features after global pooling. We compared with classic KD~\cite{hinton2015distilling}, DKD~\cite{zhao2022decoupled}, CRD~\cite{tian2019contrastive}, DIST~\cite{huang2022knowledge}, OFA~\cite{hao2024one} and LSHL2~\cite{wang2021distilling}.

\begin{table*}
    \centering
    \small
    \setlength{\tabcolsep}{2pt}
    \renewcommand{\arraystretch}{1.2}
    \begin{tabular}{lccccccccccc}
        \toprule
        \multirow{2}{*}{Dataset}                           & \multirow{2}{*}{\begin{tabular}[c]{@{}c@{}}Pretraining \\ Method\end{tabular}} & \multirow{2}{*}{Student} & \multirow{2}{*}{Teacher} &\multirow{2}{*}{Method} & \multicolumn{6}{c}{Shots per Class}                                                                                                  \\
                                                            &                                                                                &                           &  &  & 1         & 2     & 3     & 4     & 5     & 8     & 16    \\ \hline
        \hline
        \multicolumn{1}{l|}{\multirow{6}{*}{CUB}}          & \multicolumn{1}{l|}{\multirow{3}{*}{MoCov3}}                                     & \multirow{3}{*}{ViT-S} & -     & baseline   & 18.5 $\pm$ 0.7 & 27.2 $\pm$ 0.2 & 35.5 $\pm$ 1.0 & 40.0 $\pm$ 0.6 & 45.3 $\pm$ 0.9 & 54.1 $\pm$ 0.4 & 65.3 $\pm$ 0.3 \\
        \multicolumn{1}{l|}{}                              & \multicolumn{1}{l|}{}                                                          &                         & -      & IbM2   & 19.5 $\pm$ 0.4 & 27.6 $\pm$ 0.2 & 35.6 $\pm$ 0.7 & 40.2 $\pm$ 0.4 & 46.2 $\pm$ 1.1 & 55.9 $\pm$ 0.4 & 66.8 $\pm$ 0.5 \\ 
        \multicolumn{1}{l|}{}                              & \multicolumn{1}{l|}{}                                                          &                         & ViT-B      & TCS   & \textbf{20.1 $\pm$ 0.9} & \textbf{33.1 $\pm$ 0.9} & \textbf{44.5 $\pm$ 1.0} & \textbf{49.2 $\pm$ 0.5} & \textbf{55.2 $\pm$ 0.5} & \textbf{63.0 $\pm$ 0.4} & \textbf{71.8 $\pm$ 0.1} \\ \cline{2-12}
        \multicolumn{1}{l|}{}                              & \multicolumn{1}{l|}{\multirow{4}{*}{DINOv2}}                                   & \multirow{4}{*}{ViT-S} & -     & baseline & 51.0 $\pm$ 1.4
 & 66.0 $\pm$ 0.5 & 73.1 $\pm$ 0.8 & 75.1 $\pm$ 0.7 & 78.8 $\pm$ 0.5 & 82.7 $\pm$ 0.5 & 86.3 $\pm$ 0.3 \\
        \multicolumn{1}{l|}{}                              & \multicolumn{1}{l|}{}                                                          &                         & -      & IbM2 & 54.4 $\pm$ 1.0 & \textbf{66.8 $\pm$ 1.0} & 75.0 $\pm$ 0.6 & 76.9 $\pm$ 0.5 & 79.8 $\pm$ 0.5 & 83.6 $\pm$ 0.5 & 86.5 $\pm$ 0.3 \\ 
        \multicolumn{1}{l|}{}                              & \multicolumn{1}{l|}{}                                                          &                         & ViT-B      & TCS & 53.8 $\pm$ 0.5 & 64.5 $\pm$ 0.3 & 74.7 $\pm$ 0.5 & \textbf{78.0 $\pm$ 0.4} & 80.9 $\pm$ 0.6 & 84.5 $\pm$ 0.2 & \textbf{87.6 $\pm$ 0.0} \\ 
        \multicolumn{1}{l|}{}                              & \multicolumn{1}{l|}{}                                                          &                         & ViT-L      & TCS & \textbf{54.7 $\pm$ 1.7} & 65.3 $\pm$ 0.8 & \textbf{75.3 $\pm$ 1.2} &  77.8 $\pm$ 0.4 & \textbf{81.2 $\pm$ 0.5} & \textbf{84.7 $\pm$ 0.3} & 87.5 $\pm$ 0.1 \\ 
        \hline
        \multicolumn{1}{l|}{\multirow{6}{*}{ImageNet-1K}} & \multicolumn{1}{l|}{\multirow{3}{*}{MoCov3}}                                     & \multirow{3}{*}{ViT-S} & -     & baseline   & 33.1  $\pm$ 0.6 & 42.4 $\pm$ 0.3 & 47.2 $\pm$ 0.4 & 49.8 $\pm$ 0.4 & 51.6 $\pm$ 0.3 & 55.2 $\pm$ 0.2 & 59.2 $\pm$ 0.3 \\
        \multicolumn{1}{l|}{}                              & \multicolumn{1}{l|}{}                                                          &                         & -      & IbM2   & 34.4 $\pm$ 0.6 & 43.8 $\pm$ 0.3 & 48.7 $\pm$ 0.4 & 51.5 $\pm$ 0.3 & 53.0 $\pm$ 0.2 & 56.2 $\pm$ 0.1 & 60.3 $\pm$ 0.2 \\ 
        \multicolumn{1}{l|}{}                              & \multicolumn{1}{l|}{}                                                          &                         & ViT-B      & TCS   & \textbf{35.1 $\pm$ 0.4} & \textbf{46.0 $\pm$ 0.3} & \textbf{51.4 $\pm$ 0.4} & \textbf{54.2 $\pm$ 0.2} & \textbf{55.9 $\pm$ 0.3} & \textbf{58.7 $\pm$ 0.2} & \textbf{61.6 $\pm$ 0.1} \\ \cline{2-12} 
        \multicolumn{1}{l|}{}                              & \multicolumn{1}{l|}{\multirow{4}{*}{DINOv2}}                                     & \multirow{4}{*}{ViT-S} & -     & baseline   & 37.0 $\pm$ 0.7 & 46.9 $\pm$ 0.5 & 52.9 $\pm$ 0.3 & 57.0 $\pm$ 0.1 & 59.7 $\pm$ 0.4 & 64.0 $\pm$ 0.1 & 67.6 $\pm$ 0.0 \\
        \multicolumn{1}{l|}{}                              & \multicolumn{1}{l|}{}                                                          &                         & -      & IbM2   & \textbf{37.2 $\pm$ 0.9} & 47.1 $\pm$ 0.5 & 53.1 $\pm$ 0.3 & 58.6 $\pm$ 0.3 & 61.6 $\pm$ 0.2 & 65.3 $\pm$ 0.1 & 68.9 $\pm$ 0.1 \\
        \multicolumn{1}{l|}{}                              & \multicolumn{1}{l|}{}                                                          &                         & ViT-B      & TCS   & 34.2 $\pm$ 0.4 & 47.4 $\pm$ 0.6 & 54.7 $\pm$ 0.6 & 58.9 $\pm$ 0.3 & 61.3 $\pm$ 0.2 & 65.3 $\pm$ 0.4 & 69.7 $\pm$ 0.1 \\
        \multicolumn{1}{l|}{}                              & \multicolumn{1}{l|}{}                                                          &                         & ViT-L      & TCS   & 35.8 $\pm$ 0.4 & \textbf{48.2 $\pm$ 0.6} & \textbf{55.3 $\pm$ 0.5} & \textbf{59.7 $\pm$ 0.3} & \textbf{61.7 $\pm$ 0.1} & \textbf{65.6 $\pm$ 0.4} & \textbf{69.7 $\pm$ 0.1} \\ \bottomrule
    \end{tabular}
    \caption{Average of top-1 accuracy (\%) plus standard deviation across 3 random subsets on ImageNet-1K and CUB. The best result in every comparison group is highlighted in boldface.}
    \label{table:tab1_low_shot}
\end{table*}

\paragraph{Main Results} As shown in Tables~\ref{tab:cifar} and~\ref{tab:imagenet}, our TCS, especially when equipped with the eLSH module, not only achieves clearly higher accuracy than state-of-the-art methods, but also significantly reduces memory and runtime costs. The training time is almost halved, and the GPU memory footprint is reduced by roughly 25\%.

The proposed TCS is versatile to be applied upon various network architectures and sizes, and the teacher-student pair can be of heterogeneous network types. The results show that even when the capacity gap between teacher and student is large~\cite{mirzadeh2020improved}, TCS copes well with it. Furthermore, TCS requires \emph{only one} forward pass from the teacher, which explains why its training time is roughly half of other KD methods.

Compare to another teacher-free method, self-distillation by~\citet{zhang2019your}, TCS has clear advantages. Self-distillation requires significantly more parameters, larger GPU memory and computing time, but its accuracy lags behind TCS consistently.


\subsection{Experiments in few-shot learning}

Consistent with the protocols outlined in the pFSL (practical few-shot learning) paper~\cite{fu2024ibm2}, we freeze the backbones of both the teacher and student networks during the training phase. Despite the recognized challenges associated with distilling knowledge from teacher models trained via unsupervised or self-supervised methods, our approach effectively overcomes these hurdles. In our experiments, we exclusively employed self-supervised models as teachers, demonstrating our method's capability to extract and distill valuable information from these models. 

We experimented on two datasets: ImageNet-1K~\cite{russakovsky2015imagenet} and CUB-200-2011 (CUB)~\cite{wah2011caltech}. CUB has 11,788 images across 200 bird species. For both datasets, we randomly sampled the training sets following the pFSL protocol by selecting $k$ images per class (i.e., $k$-shot learning) and formed the novel set with all classes (1000 for ImageNet, 200 for CUB). 

The evaluation was on the full validation set. This setting is different from classic FSL, where the number of classes is very small (mostly 5, in comparison to 1000 or 200). pFSL argues that being able to handle a large number of classes is essential for making few-shot learning practical.

Experiments were conducted using two types of backbones: Vision Transformer (ViT)~\cite{dosovitskiy2020image} and ResNet50~\cite{he2016deep}. The teacher network's pretraining methods included DINOv2~\cite{oquab2023dinov2} and MoCov3~\cite{chen2021empirical}. We mainly compare our TCS method with the IbM2 approach proposed by~\citet{fu2024ibm2}.

\paragraph{Main Results} Results in Table~\ref{table:tab1_low_shot} show that our TCS method improves few-shot learning by comparing the top-1 accuracy of TCS versus IbM2. TCS achieves top accuracy in almost all cases, except in two experiments with either 1- or 2-shot learning. As the number of training shots $k$ increases, TCS's margin grows to about 1.0\% and can exceed 2\% in some cases. Another benefit of our TCS is that it is robust---except for when $k=1,2$, the standard deviation of TCS is often small.

This trend can be attributed to the PCA computation. When $k$ is too small ($k=1,2$), the number of examples used to learn PCA coefficients may even be smaller than its dimensionality. Hence, the coordinate system defined by PCA is low-quality in such scenarios, which leads to relatively worse TCS results.

Notably, unlike previous studies that highlight sensitivity to learning rate variations in few-shot learning scenarios~\cite{hu2022pushing, fu2024ibm2}, our method demonstrates robust performance with a fixed learning rate, thus reducing the necessity for extensive hyperparameter adjustments.

Finally, as shown in Table~\ref{table:tab1_low_shot}, different pretraining methods for the teacher (MoCov3 or DINOv2) and different teachers (ViT-B or ViT-L) all lead to robust TCS results. A larger teacher with a better pretraining (ViT-L plus DINOv2) in fact lead to the best set of results.

\subsection{Ablation studies}

We initially investigate the impact of model size discrepancies on teacher-student relationships and the effects of the coordinate system.

\paragraph{Impact of model size: Teacher vs. Student} Substantial differences in model size between the teacher and student networks can often be counterproductive, i.e., large capacity gap is harmful~\cite{mirzadeh2020improved} in existing KD methods. In contrast, our TCS shows resilience to these discrepancies and even demonstrates improvements when the teacher is small but the student is large. We hypothesize that so long as there is large enough difference between the teacher and student, TCS captures useful dark knowledge from the teacher and helps the student---because the student mainly inherits the coordinate system, not the details of the teacher network.

Detailed results are provided in Table~\ref{tab:model-size} for pFSL, with experiments conducted on CUB 16-shot and ImageNet-1K 16-shot. As these results suggest, large capacity gap is not harmful to TCS: All teachers (small or large) are helpful, leading to accuracy higher than the baseline (whose teacher `T' is marked as `-'). Furthermore, when the teacher is smaller (ViT-S) but student is larger (ViT-B), TCS even can achieve 2-4\% accuracy increase.

\begin{table}
	\centering
	\small
	\setlength{\tabcolsep}{4pt} 
	\renewcommand{\arraystretch}{1.2} 
	\begin{tabular}{lcccc}
	\toprule
	\multirow{2}{*}{Dataset} & \multirow{2}{*}{\begin{tabular}[c]{@{}c@{}}Pretraining \\ Method\end{tabular}} & \multirow{2}{*}{S} & \multirow{2}{*}{T} & \multirow{2}{*}{Top 1 (\%)} \\ 
	& & & & \\ 
	\hline
	\multirow{4}{*}{ImageNet-1K} & \multirow{4}{*}{MoCov3} & \multirow{2}{*}{ViT-S} & - & $59.30 \pm 0.27$ \\
							  &                       &                          & ViT-B & $61.59 \pm 0.12$ \\
							  \cline{3-5}
							  &                       &\multirow{2}{*}{ViT-B} & - & $63.76 \pm 0.25$ \\
							  &                       &                          & ViT-S & $65.73 \pm 0.05$ \\
	\hline
	\multirow{4}{*}{CUB} & \multirow{4}{*}{MoCov3} & \multirow{2}{*}{ViT-S} & - & $65.58 \pm 0.30$  \\
						 &                       &                           & ViT-B & $71.80 \pm 0.05$  \\
						 \cline{3-5}
						 &                       & \multirow{2}{*}{ViT-B} & - & $67.32 \pm 0.21$ \\
						 &                       &                           & ViT-S &  $71.17 \pm 0.26$\\
	
	\hline
	\multirow{5}{*}{CUB} & \multirow{5}{*}{DINOv2} & \multirow{3}{*}{ViT-S} & - & $87.00 \pm 0.20$ \\
						 &                         &                           & ViT-B & $87.47 \pm 0.09$ \\
						 &                         &                           & ViT-L & $87.52 \pm 0.06$ \\
						 \cline{3-5}
						 &                         & \multirow{2}{*}{ViT-B} & - & $88.71 \pm 0.19$ \\
						 &                       &                           & ViT-S &  $88.70 \pm 0.22$\\
	\bottomrule
	\end{tabular}
	\caption{Impact of model size and capacity gap. Top-1 accuracy (\%) along with the standard deviation across three random subsets on ImageNet-1K and CUB are reported.}
	\label{tab:model-size}
\end{table}	

\paragraph{Effect of the coordinate system} The dark knowledge in TCS resides in the coordinate system computed from PCA. In this study, we replace it with $D_t$ random orthogonal vectors (`\emph{random}' in Table~\ref{tab:effect-E}). We also study using different data to compute PCA. The first group (`In-domain') used training data from the target task with 3- or 16-shot or all the training data to compute PCA. `N/A' means TCS is not used. The second group (`Out-of-domain') used either random PCA coefficients or computed PCA using 16-shot data \emph{not} in the target task. For example, in the last row of Table~\ref{tab:effect-E}, PCA was computed using the `Diabetic Retinopathy' dataset although the target task is CUB.

As shown in Table~\ref{tab:effect-E}, it is not surprising that random PCA coefficients is not helpful. But, even the coordinate system computed from out-of-domain data consistently improves the accuracy by roughly 2\% compared to the baseline not using TCS. It is also not surprising that in-domain PCA is better than out-of-domain PCA by 3-4\%. The more data used for computing the coordinate system, the higher the accuracy. These results again verifies that the coordinate system indeed contains essential dark knowledge, or, as we argued, ``all you need is a tailored coordinate system'' in KD.

\begin{table}
\centering
\small
\setlength{\tabcolsep}{3pt}
\begin{tabular}{ccc}
\toprule
  & Data for PCA  & Top 1 (\%) \\ 
\midrule
  \multirow{4}{*}{In-domain}  & N/A & $65.30 \pm 0.30$ \\
   & 3-shot & $70.37 \pm 0.11$ \\
   & 16-shot & $71.80 \pm 0.05$ \\
   & all & $71.90 \pm 0.29$ \\
  \midrule
 \multirow{4}{*}{Out-of-domain} & \emph{random}  & $65.71 \pm 0.61$ \\
  & Flowers102  & $67.56 \pm 0.59$ \\
  & Pets  & $67.12 \pm 0.66$ \\
  & Diabetic Retinopathy  & $67.54 \pm 0.43$ \\
\bottomrule
\end{tabular}
\caption{Results of computing the PCA coordinate system using different data. Experiments were on the CUB dataset for 16-shot pFSL. The three datasets used in the out-of-domain category all come from VTAB-1K~\cite{vtab}.}
\label{tab:effect-E}
\end{table}

\section{Conclusion}

We targeted the knowledge distillation (KD) task from a perspective very different from existing methods. In this paper, we argued that the dark knowledge in a teacher network can be captured by knowing its coordinate system (the linear subspace where the teacher's features reside in), and by tailoring this coordinate system to the target task, the student can be effectively injected the dark knowledge. Hence our method is named TCS.

In TCS, the coordinate system is found by PCA, and tailored to the target task via a proposed iterative feature selection method. TCS was successfully applied in KD and practical few-shot learning. TCS is teacher-free and applies to diverse architectures, allows cross-architecture distillation with large capacity gap. It achieves significantly higher accuracy than state-of-the-art KD methods, while only requiring roughly half of their training time and GPU memory costs.

TCS has its limitations. In this paper, it was tested in KD (classification). Given the versatility of TCS, we plan to extend this method to distillation of object detection and segmentation models. Also, since TCS works well for teacher networks learned through self-supervised learning, it is also interesting and promising to extend TCS to compressing and accelerating large language models (LLMs) or large multimodal models.

In the future, we are also interested in its theoretical foundation: how is the dark knowledge encoded by the coordinate system?

\section{Acknowledgments}
This research was partly supported by the National Natural Science Foundation of China under Grant 62276123.

\bibliography{aaai25}

\end{document}